\documentclass{article} % For LaTeX2e
\usepackage{iclr2021_conference,times}
\usepackage{multirow}

\usepackage{amsmath,amsfonts,bm}

\def\eqref#1{equation~\ref{#1}}
\def\1{\bm{1}}

\DeclareMathAlphabet{\mathsfit}{\encodingdefault}{\sfdefault}{m}{sl}
\SetMathAlphabet{\mathsfit}{bold}{\encodingdefault}{\sfdefault}{bx}{n}

\usepackage{hyperref}
\usepackage{url}

\usepackage{inconsolata}

\usepackage{times}
\usepackage{latexsym}
\usepackage{amsfonts}
\usepackage{amsmath}
\usepackage{amssymb}
\usepackage{multicol}
\usepackage{multirow}
\usepackage{xspace}
\usepackage{booktabs}
\usepackage{bbding}
\usepackage{array}
\usepackage{threeparttable}
\usepackage{tcolorbox}
\usepackage{tabularx}
\usepackage{enumitem}
\usepackage[linesnumbered,ruled,vlined]{algorithm2e}
\usepackage{xcolor,colortbl}
\usepackage{setspace}
\usepackage{makecell}
\usepackage{xltabular}
\usepackage{graphicx}
\usepackage{subfig}

\usepackage{cleveref}
\crefformat{section}{\S#2#1#3}
\crefformat{subsection}{\S#2#1#3}
\crefformat{subsubsection}{\S#2#1#3}
\crefrangeformat{section}{\S\S#3#1#4 to~#5#2#6}
\crefmultiformat{section}{\S\S#2#1#3}{ and~#2#1#3}{, #2#1#3}{ and~#2#1#3}
\crefmultiformat{subsection}{\S\S#2#1#3}{ and~#2#1#3}{, #2#1#3}{ and~#2#1#3}
\Crefformat{figure}{#2Fig.~#1#3}
\Crefmultiformat{figure}{Figs.~#2#1#3}{ and~#2#1#3}{, #2#1#3}{ and~#2#1#3}
\Crefformat{table}{#2Tab.~#1#3}
\Crefmultiformat{table}{Tabs.~#2#1#3}{ and~#2#1#3}{, #2#1#3}{ and~#2#1#3}
\Crefformat{appendix}{Appx.~\S#2#1#3}
\crefmultiformat{appendix}{Appx.~\S#2#1#3}{ and~#2#1#3}{, #2#1#3}{ and~#2#1#3}
\crefformat{algorithm}{Alg.~#2#1#3}
\Crefformat{equation}{Eq.~#2#1#3}

\newcommand{\xhdr}[1]{\vspace{0.3em}\noindent{{\bf #1.}}}
\newcommand{\modelname}{\textsc{InsTag}\xspace}

\newcommand{\lmname}{\textsc{TagLM}\xspace}

\title{\#InsTag: Instruction Tagging for Analyzing Supervised Fine-tuning of Large Language Models}

\author{Keming Lu\thanks{Equally Contributed. Order determined by swapping the one in \citet{yuan-etal-2023-exploring}.} $\ $\& Hongyi Yuan$^*$\thanks{Work done during internships at Alibaba DAMO Academy.} \& Zheng Yuan \& Runji Lin$^\dagger$  \\
Alibaba DAMO Academy \\
\texttt{\{lukeming.lkm,yuanhongyi.yhy,yuanzheng.yuanzhen,linrunji.lrj\}@alibaba-inc.com} \\
\AND
Junyang Lin \& Chuanqi Tan \& Chang Zhou \& Jingren Zhou \\
Alibaba DAMO Academy \\
\texttt{\{junyang.ljy,chuanqi.tcq,ericzhou.zc,jingren.zhou\}@alibaba-inc.com} \\
}

\usepackage{wrapfig,lipsum}
\begin{document}

\maketitle

\begin{abstract}
Foundation language models obtain the instruction-following ability through supervised fine-tuning (SFT).
Diversity and complexity are considered critical factors of a successful SFT dataset, while their definitions remain obscure and lack quantitative analyses.
In this work, we propose \modelname, an open-set fine-grained tagger, to tag samples within SFT datasets based on semantics and intentions and define instruction diversity and complexity regarding tags.
We obtain 6.6K tags to describe comprehensive user queries.
We analyze popular open-sourced SFT datasets and find that the model ability grows with more diverse and complex data.
Based on this observation, we propose a data selector based on \modelname to select 6K diverse and complex samples from open-source datasets and fine-tune models on \modelname-selected data.
The resulting models, \lmname, outperform open-source models based on considerably larger SFT data evaluated by \textsc{MT-Bench}, echoing the importance of query diversity and complexity.
We open-source \modelname in \url{https://github.com/OFA-Sys/InsTag}.
\end{abstract}

\section{Introduction}
The rise of contemporary chatbots, including \textsc{GPT-4} \citep{openai2023gpt4}, has brought to the forefront of generative artificial intelligence that is based on large language models~(LLMs) to tackle a variety of real-world tasks. 
Well-aligned LLMs with human expectations can precisely recognize human intentions and properly formalize responses expressed in natural languages \citep{wang2023aligning}.
Achieving such a level of alignment typically necessitates fine-tuning processes, such as supervised fine-tuning~(SFT) \citep{alpaca,vicuna2023,llama2},  response ranking \citep{yuan2023rrhf,song2023preference,rafailov2023direct}, and reinforcement learning with human feedback~(RLHF) \citep{bai2022training,instructgpt,llama2}, to enable LLMs to comprehend and execute diverse instructions effectively.

A broad range of training data covering various semantics and specialties is crucial for achieving alignment with human preference through SFT, which is typically gathered through crowd-sourcing \citep{instructgpt,bai2022training,llama2} or by distilling from other LLMs \citep{alpaca,ding2023enhancing}. 
The SFT data for alignment is generally formalized in a multi-turn utterance manner, and each turn is composed of a human query and a corresponding response expected to generate by well-aligned chatbots. 
Recent research indicates that the training dataset for alignment should be diverse and complex, covering various domains, tasks, and formats \citep{xu2023wizardlm,mukherjee2023orca,wang2023far}. 
Such diversity and complexity are mainly determined by query formation. 
Various methods are proposed and claimed to improve the diversity and complexity of the queries and advance the alignment of LLMs (\citealt{wang2023selfinstruct,xu2023wizardlm,ding2023enhancing}; \textit{inter alia}). 
However, how to quantify the diversity and complexity of queries is significantly understudied and hence less analyzed.

To shed light on this topic, we propose using a tagging system to feature and categorize samples in SFT datasets. 
Given the versatile tasks aligned LLMs are expected to handle, an equally versatile tag system is necessary to distinguish open-world human queries.
However, building an open, fine-grained tagging system manually is infeasible to scale for large datasets.
To this end, we propose \modelname, an automatic \textsc{Ins}truction \textsc{Tag}ging method empowered by proprietary high-performing chatbot \textsc{ChatGPT}\footnote{\url{https://chatgpt.openai.com}}. 
Leveraging such a well-aligned chatbot, \modelname designs a framework to prompt \textsc{ChatGPT} to automatically assigns tags to queries. 
\modelname achieves the increased quality of the tag assignment by deliberately prompting \textsc{ChatGPT} to explain each tag assigned and including a systematic tag normalization procedure.  
We apply \modelname to a collection of existing rich open-source SFT datasets and build open-set, fine-grained tags which, as we observed, can reflect the semantics and intentions beneath human queries in SFT datasets. 
Through the scope of tags, we conduct a detailed and quantified analysis of existing open-source datasets, providing insights into query distributions in terms of diversity and complexity.
Such analyses reveal that diverse and complex queries induce high alignment performance during SFT.
Following this insight, we propose a data selector based on \modelname, including a complexity-focus diverse sampling method that can extract the most complex queries in a diverse distribution.
LLMs fine-tuned with data selected by the \modelname selector perform well on the popular benchmark \textsc{MT-Bench}\citep{zheng2023judging}, supporting our previous query distribution insights.

% Contributions
The contributions of this work are mainly three-fold.
Firstly, we propose using open-set fine-grained intention tags as instruction diversity and complexity metrics. To this end, we develop \modelname, an annotator that leverages the instruction-following abilities of proprietary chatbots and employs tag normalization methods.
Secondly, we analyze existing open-source SFT datasets and provide insights into query diversity and complexity.
Finally, we design a data selector based on \modelname and apply it to the latest open-source datasets. The resulting best LLMs, \lmname-13b-v1.0 and \lmname-13b-v2.0 respectively based on LLaMA \citep{touvron2023llama} and LLaMA-2 \citep{llama2}, aligned with selected data achieve scores of 6.44 and 6.55 on the benchmark \textsc{MT-Bench}, respectively, surpassing a group of LLMs aligned with considerably more SFT data. 
Our contributions are verified
with experiments and multifaceted analysis.
Most notably, \modelname exhibits its robust potential to offer deeper insights into aligning LLMs, extending beyond the data selection demonstrated in our work.

\section{Related Works}
\xhdr{LLMs for Human Alignment}
Through supervised fine-tuning (SFT), response ranking, or reinforcement learning \citep{instructgpt,bai2022training,bai2022constitutional,yuan2023rrhf,rafailov2023direct,song2023preference,llama2}, LLMs can obtain versatile abilities for understanding and following diversified human queries expressed in natural languages, aligning with human intentions. 
Recent research mainly focused on SFT to align LLMs with human intentions and has contributed essential practices to developing open-resourced well-aligned LLMs, which is adequately summarized by \cite{zhao2023survey}. 
Several prominent works collected SFT data through human annotated demonstrations \citep{instructgpt,llama2}, online user logs of proprietary LLMs \citep{vicuna2023,openchat,2023openassistant}, or prompting proprietary high-performing LLMs such as \textsc{ChatGPT} or \textsc{GPT-4} \citep{openai2023gpt4} to generate or rewrite samples (\citealt{alpaca,ding2023enhancing,xu2023wizardlm,mukherjee2023orca}; \textit{inter alia}). 
Different LLMs fine-tuned on the datasets have aligned with human preference and exhibited good performance in various real-world tasks.

\xhdr{Data for Human Alignment}
It has been highlighted that the performance of aligned LLMs is affected by the quality of the SFT data. Such data quality pertains to the level of responses \citep{peng2023instruction,vicuna2023}, the difficulty of tasks presented \citep{mukherjee2023orca}, the complexity of queries \citep{xu2023wizardlm}, the diversity of semantics \citep{ding2023enhancing,alpaca}, and the scale of sample amounts \citep{zhou2023lima}. 
\citet{alpaca} used Self-Instruct \citep{wang2023selfinstruct} to generate diversified queries for SFT and \citet{xu2023wizardlm} proposed Evol-Instruct to complexify simple queries for better human alignment. 
\citet{mukherjee2023orca} used proprietary high-performing LLMs to rewrite the queries and responses of samples from FLAN collection \citep{longpre2023flan} and observed improvement of LLMs in conventional NLP task solving.  
\citet{ding2023enhancing} proposed UltraChat using manually designed diverse anchor concepts and entities to generate multi-turn data by inducing conversations in \textsc{ChatGPT}. 
OpenChat \citep{openchat} and Vicuna \citep{vicuna2023} are both current open-sourced LLMs with cutting-edge instruction following abilities, and both models are trained on the user logs of \textsc{GPT-4} from ShareGPT \footnote{\url{https://sharegpt.com/}}. 
As evaluated in \citet{wang2023far}, the success of fine-tuning on ShareGPT demonstrates that queries from user logs are of higher diversity and the responses generated from \textsc{GPT-4} are of better quality, resulting in superior instruction following the abilities. 
\citet{zhou2023lima} found that a small amount of high-quality data is sufficient for LLMs to excel at human alignment. 

Although current research proposed more diversified and complexified SFT data and made significant progress in developing well-aligned LLMs with human intentions, existing works have yet to discuss how to quantify the diversity and complexity of queries.
Taking advantage of the high-performing \textsc{ChatGPT} and \textsc{GPT-4}, we annotate existing data samples with tag entities.
Through the scope of tags, we quantify the diversity and complexity of the training data for the first time and study the data mixture for better alignment.

\begin{figure}[t]
    \centering
    \includegraphics[width=\linewidth]{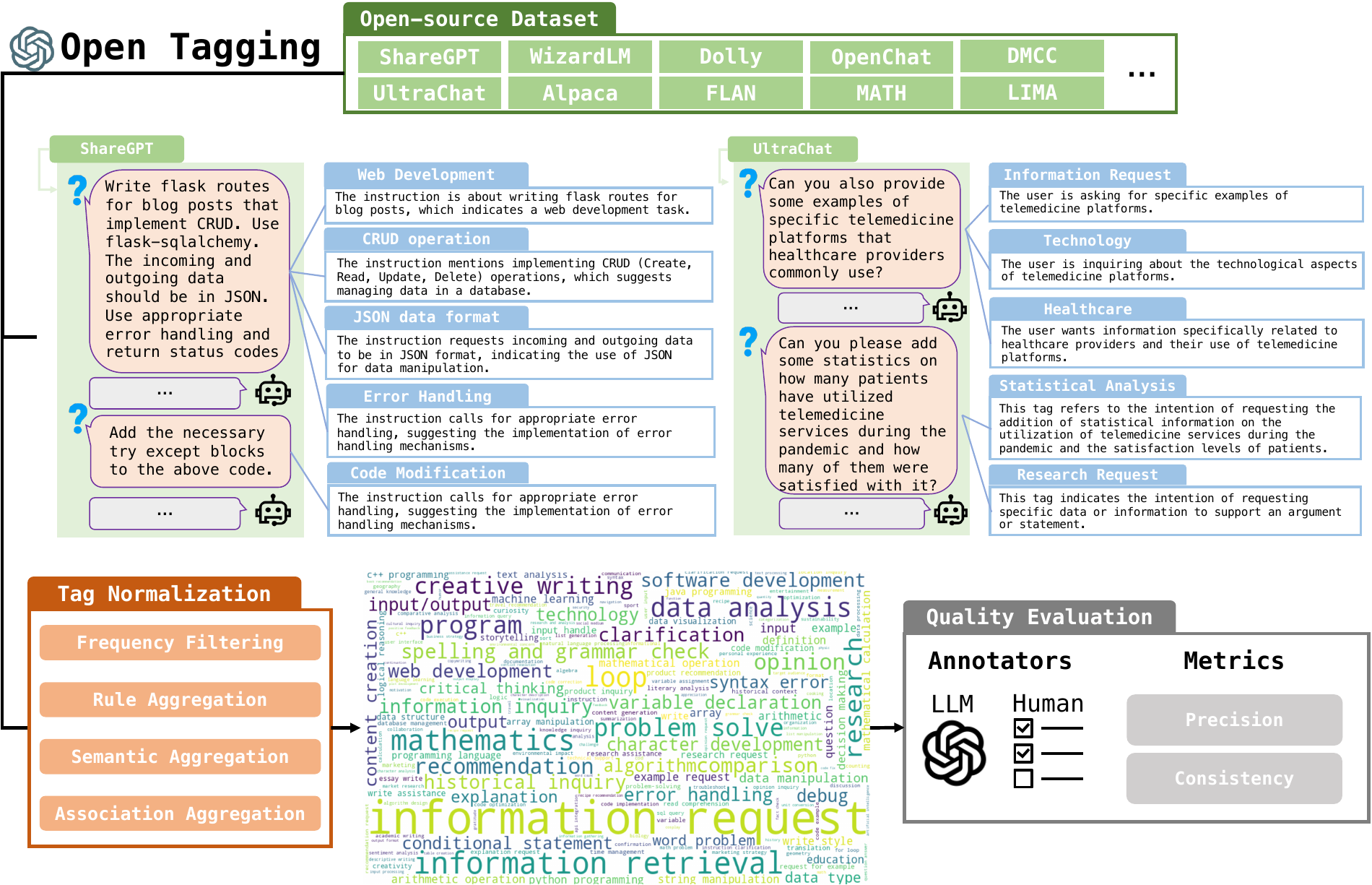}
    \caption{
        Overview of \modelname.
        We use \textsc{ChatGPT} to annotate fine-grained intention tags for a series of open-source datasets.
        We present two cases of open tagging annotations in this figure, selected from ShareGPT and UltraChat.
        A tag normalization, including multiple denoising and aggregation methods, is then applied to the original tagging results.
        Finally, the quality of the tag set, as shown in the word cloud, is evaluated by human and LLM annotators, focusing on the tagging precision and consistency.
    }
    \label{fig:main}
\end{figure}

\section{InsTag}
This section presents an automatic instruction tagging method to identify proper tags describing instruction intentions in an open setting.
We first define fine-grained intention tags and present the open tagging process with LLMs~(\Cref{sec:definition}).
Then, we design a systematic normalization method to denoise the raw tags from previous annotations~(\Cref{sec:normalization}).
We also fully evaluate the tagging quality to ensure \modelname generates precise and consistent intention tags~(\Cref{sec:quality_eval}).
Finally, we use \modelname to analyze open-source SFT datsets~(\Cref{sec:sft_data_analysis}).

\subsection{Open-set Fine-grained Tagging}\label{sec:definition}
Instructions, or queries in the context of modern chatbots, serve as expressions of user intentions, which can often be multifaceted and highly intricate.
To illustrate, we showcase an instruction from the ShareGPT dataset in \Cref{fig:main}, where the user submits a coding request specifying desired output formats and error-handling methods.
To better parse such instructions, it is essential to employ fine-grained tags that can identify atomic intentions rather than relying on generalized, coarse-grained classes.
However, although fine-grained intention tags offer a more detailed understanding of instruction distribution, they also present challenges in annotation and normalization.
Therefore, we propose an open-set tagging with \textsc{ChatGPT} and a normalization technique to address these issues.
In other words, we do not provide a predefined ontology of tags when annotating with \textsc{ChatGPT}.
We choose an open setting since a closed set is not flexible enough to cover versatile intentions in open chatting.

We use the prompt shown in \Cref{tab:tagging_prompt} to employ \textsc{ChatGPT}, providing fine-grained intention tags for given queries.
We provide few-shot examples in the prompt to hint \textsc{ChatGPT} provides tags in a specific format and requires \textsc{ChatGPT} to provide JSON outputs for parsing.
As shown in \Cref{fig:main}, we separately annotate each query in a chat session and require \textsc{ChatGPT} to briefly explain tags for the convenience of quality evaluation.
The number of original tags annotated by \textsc{ChatGPT} is larger than 12 thousand, showing \textsc{ChatGPT} can provide diverse and fine-grained annotations.
However, we notice the original tagging results provided by \textsc{ChatGPT} contain noticeable noises, including inconsistency in word format and granularity.
Therefore, we design a systematic method to normalize the open-set tagging results from \textsc{ChatGPT}.

\subsection{Tag Normalization}\label{sec:normalization}

\begin{table}[t]
    \centering
    \small
    \vspace{-3pt}
    \setlength{\tabcolsep}{2mm}{
    \begin{tabular}{p{3.3cm}p{6cm}p{3cm}}
    \toprule
    \textbf{Inconsistency} & \textbf{Examples} &  \textbf{Output} \\
    \midrule
    \makecell[l]{Lexical Noise\\ \textbf{Rule Aggregation}} & Information Retrieval, information\_retrieval, infomation retrieve  & information retrieval \\
    \midrule
    \makecell[l]{Uncontrolled Granularity\\ \textbf{Semantic Aggregation}} & information request, request for information, request for additional information, request for more information, additional information request, specific information request & information request\\
    \midrule
    \makecell[l]{Spurious Correlations\\ \textbf{Association Aggregation}} & (mathematics, math problem), (loop, for loop) & mathematics, for loop \\
    \bottomrule
    \end{tabular}}
    \caption{
    Inconsistency in intention tagging results from open-set annotations.
    Inconsistencies can be addressed with three aggregation methods described in \Cref{sec:normalization}.
    }
    \label{tab:inconsistency}
    \vspace{-3pt}
\end{table}

The production of intention tags through \textsc{ChatGPT} in an open setting presents a challenge in ensuring consistency, as no predefined ontology is provided, resulting in noise in the raw tagging outcomes.
We have identified three significant types of noise, detailed in \Cref{tab:inconsistency}, which have the potential to impact downstream processes:
\textbf{Lexical Noise}, arises from the instability of \textsc{ChatGPT} in adhering to output format instructions and can be mitigated through the use of stemming as a post-processing step;
\textbf{Uncontrolled Granularity} refers to the potential for \textsc{ChatGPT} to produce tags that are overly specific, such as ``request for more information'' as shown in \Cref{tab:inconsistency};
\textbf{Spurious Correlations} refer to tags that often appear together due to the bias of \textsc{ChatGPT} or data distributions.
Such tag groups should be merged to form an atomic tag.
These issues must be addressed to ensure that intentions are accurately identified and utilized in downstream processes.
Therefore, we normalize open-set tagging results by various aspects, including frequency, format, semantics, and associations.
Specifically, we clean the raw tagging results with the following normalization procedure:
\begin{itemize}[leftmargin=1em]
    \setlength\itemsep{-0.1em}
    \item \textbf{Frequency Filtering}: We first filter out long-tail tags appearing less than $\alpha$ times in the whole annotated dataset.
    $\alpha$ is a hyperparameter related to the scale of the dataset.
    \item \textbf{Rule Aggregation}: We transform all tags into lower characters to avoid the influence of capitalization. We also replace all special characters into spaces to further aggregate the tags. Finally, we apply stemming to each tag with the support of NLTK~\citep{bird2009natural}.
    \item \textbf{Semantic Aggregation}: We employ text embedding models to obtain the semantics of tags.
    We use \textsc{PhraseBERT}~\citep{wang2021phrase}, a BERT-based model designed explicitly for embedding phrases, such as titles of tags.
    Other embedding methods, such as OpenAI embeddings or \textsc{DensePhrase}~\citep{lee2020learning}, can also be adopted as alternatives.
    After we obtain the semantic embeddings of tags, we use \textsc{DBSCAN} algorithm~\citep{hahsler2019dbscan} to cluster tags with a given threshold $t$ of semantic similarity.
    Similarly, other density clustering methods can be used instead of DBSCAN for the same denoising purpose.
    Semantic aggregation controls the granularity of tags in terms of semantic similarities.
    \item \textbf{Association Aggregation}: We notice \textsc{ChatGPT} tends to provide highly associated tags that are expected to consider as an atomic tag as a whole, which mainly occurs in mathematics and coding queries.
    Therefore, we analyze all raw tagging results and employ the FP-Growth algorithm~\citep{han2000mining} to mine association rules between tags.
    We then recursively merge associated tags based on the above association rules and reduce verbosity.
\end{itemize}

We apply \modelname on widely-used open-source SFT datasets
introduced in \Cref{sec:datasets} with details.
Over 100 thousand original unique tags are generated following the \textsc{ChatGPT} annotation.
To filter out long-tail cases, we implement Frequency Filtering with a value of $\alpha=20$, resulting in the retention of 8,541 entities.
We apply the rule aggregation to address lexical noise, which merged tags and reduced the count to 7,157. 
We then utilize semantic aggregation, implementing DBSCAN clustering with a minimum semantic similarity of $0.05$, to further merge and reduce the count to 6,587 tags.
Finally, we employed the association aggregation with a minimum support of 40 times and a minimum confidence of 99\%, producing 1,772 association rules to transform tag groups into atomic tags. 
These measures were essential in streamlining the tagging process and ensuring the quality of downstream processes.

An overview of these six thousand tags locates in \Cref{app:sunburst}.
We also train a local specialized tagging LLM, \textsc{InsTagger}, based on normalized tagging data to distill such annotation abilities into smaller LLMs, shown in \Cref{app:tagger}.

\begin{table}[t]
    \centering
    \small
    \setlength{\tabcolsep}{1mm}{
    \begin{tabular}{lcccc}
    \toprule
    \multirow{2}{*}{\textbf{Metric}} & \multirow{2}{*}{\textbf{GPT-4 Annotation}} & \multirow{2}{*}{\textbf{Human Annotation~(1\%)}} & \multicolumn{2}{c}{\textbf{Agreement~($\kappa$)}}\\
    &  &  & Human-Human & Human-GPT \\
    \midrule
    \textbf{Tag Precision} & 96.1 & 100 & 0.47 & 0.92\\
    \textbf{Tag Consistency} & 86.6 & 100 & 0.73 & 0.75\\
    \bottomrule
    \end{tabular}}
    \caption{
    Evaluation for the tagging quality of \modelname.
    We design two metrics, tagging precision and consistency, for evaluating \modelname.
    We employ \textsc{GPT-4} to label 4,000 tagging results.
    Moreover, we also employ three human annotators to annotate 1\% cases and report their \underline{majority voting}.
    We report agreement between human annotators in Fleiss-kappa scores and agreement between majority voting and \textsc{GPT-4} in Cohen's kappa scores.
    We also create counterfactual cases to probe the judgment abilities of different annotators shown in \Cref{tab:tagging_evaluation_counterfactual}.
    }
    \label{tab:tagging_evaluation}
\end{table}

\subsection{Quality Evaluation}\label{sec:quality_eval}
We employ both \textsc{GPT-4} and human annotators to provide judgments in a set of randomly sampled tagging results.
We evaluate the quality of the normalized tagging dataset in precision and consistency:

\xhdr{Precision}
We define precision as whether tags assigned to a specific query are all correctly related to query intentions.
Tag precision is essential since fine-grained tags should be precisely expressed as part of query intentions.
For example, given a case $(q,\mathcal{T})$ where $q$ is the query and $\mathcal{T}$ is tags assigned to it, we employ annotators to identify any incorrect tags in $\mathcal{T}$.
We consider it a negative case if any tag in $\mathcal{T}$ is annotated as incorrect. Otherwise, it is a precise tagging case.

\xhdr{Consistency}
To form a consistent tag ontology, we naturally require that the semantics of a specific tag will not shift across queries.
An annotation case in consistency $(t, \mathcal{I})$ contains a tag $t$ and a set of randomly selected instructions $\mathcal{I}$ annotated with such tag.
Annotators are required to identify any semantic changes in tags across all instructions.

To be more specific, we randomly sample 4,000 cases, 2,000 each for precision and consistency, for \textsc{GPT-4} annotation.
Then, we hire three annotators to manually label 40 cases (1\%) selected from the above evaluation set.
Annotation from human annotators provides judgments and reveals confidence of \textsc{GPT-4} annotation on a larger scale.

The evaluation results are shown in \Cref{tab:tagging_evaluation}.
\textsc{GPT-4} provides 96.1\% and 86.6\% accuracy in tag precision and consistency, respectively.
Meanwhile, we also report the majority voting of human annotators, which suggests a hundred-percent correctness among both tasks.
We notice the Fleiss-kappa between human annotators reaches the basic agreement. In contrast, Cohen's kappa between majority voting and \textsc{GPT-4} reaches more than 0.7, suggesting a solid agreement between human and \textsc{GPT-4} annotators.
To eliminate the possibility that such results contain robust false positive annotations, we specifically design counterfactual annotation experiments shown in \Cref{tab:tagging_evaluation_counterfactual} and proof that both human and \textsc{GPT-4} are capable of precisely recalling incorrect cases.
Therefore, tags provided by \modelname are of good quality regarding precision and consistency for downstream analyses.

\begin{figure}[t]
    \centering
    \subfloat[Diversity and Complexity based on Tags]{
        \includegraphics[width=0.48\textwidth]{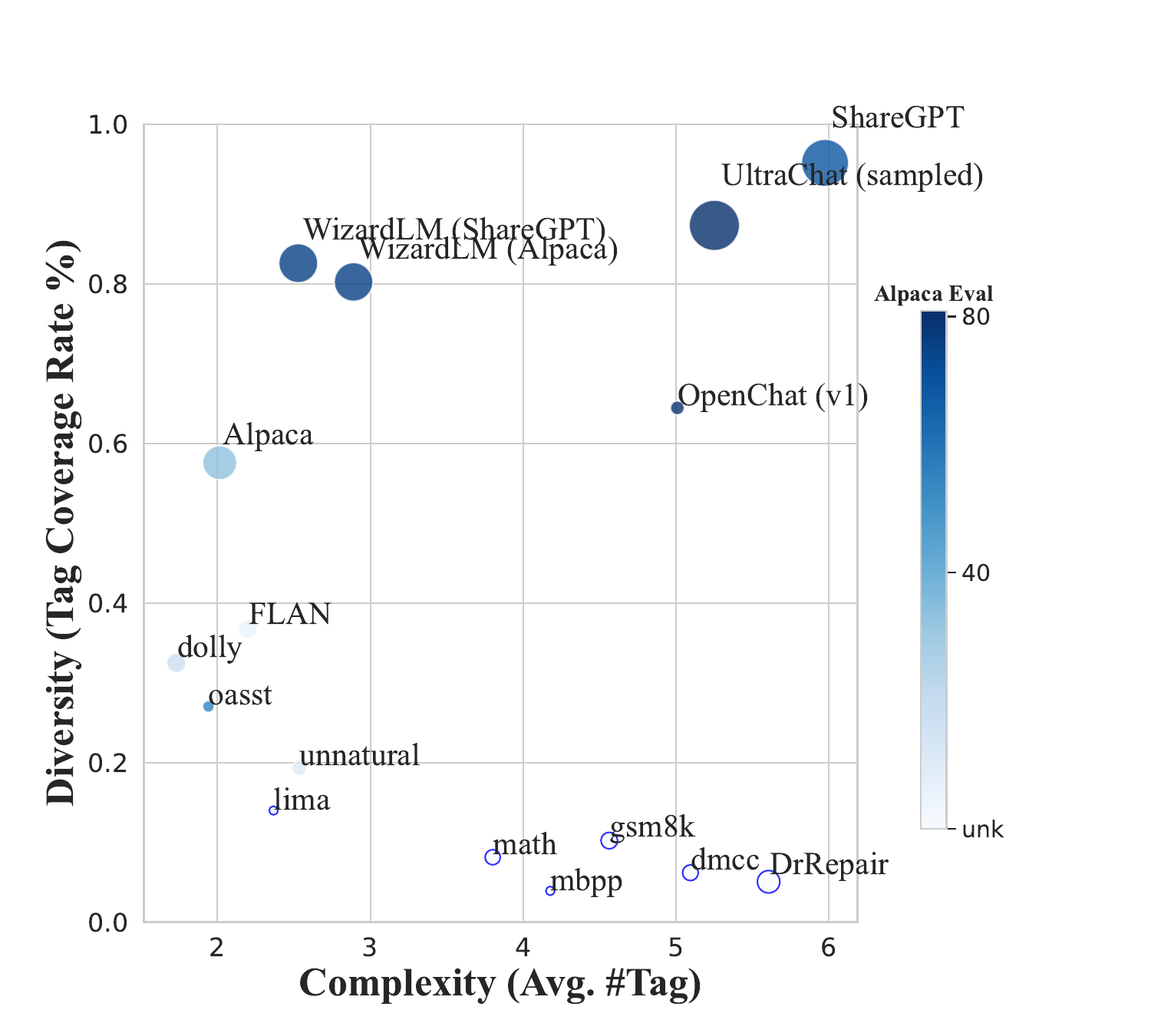}
        \label{subfig:diversity_and_complexity}
    }
    \subfloat[Dataset Correlation (Column recalls Row)]{
        \includegraphics[width=0.48\textwidth]{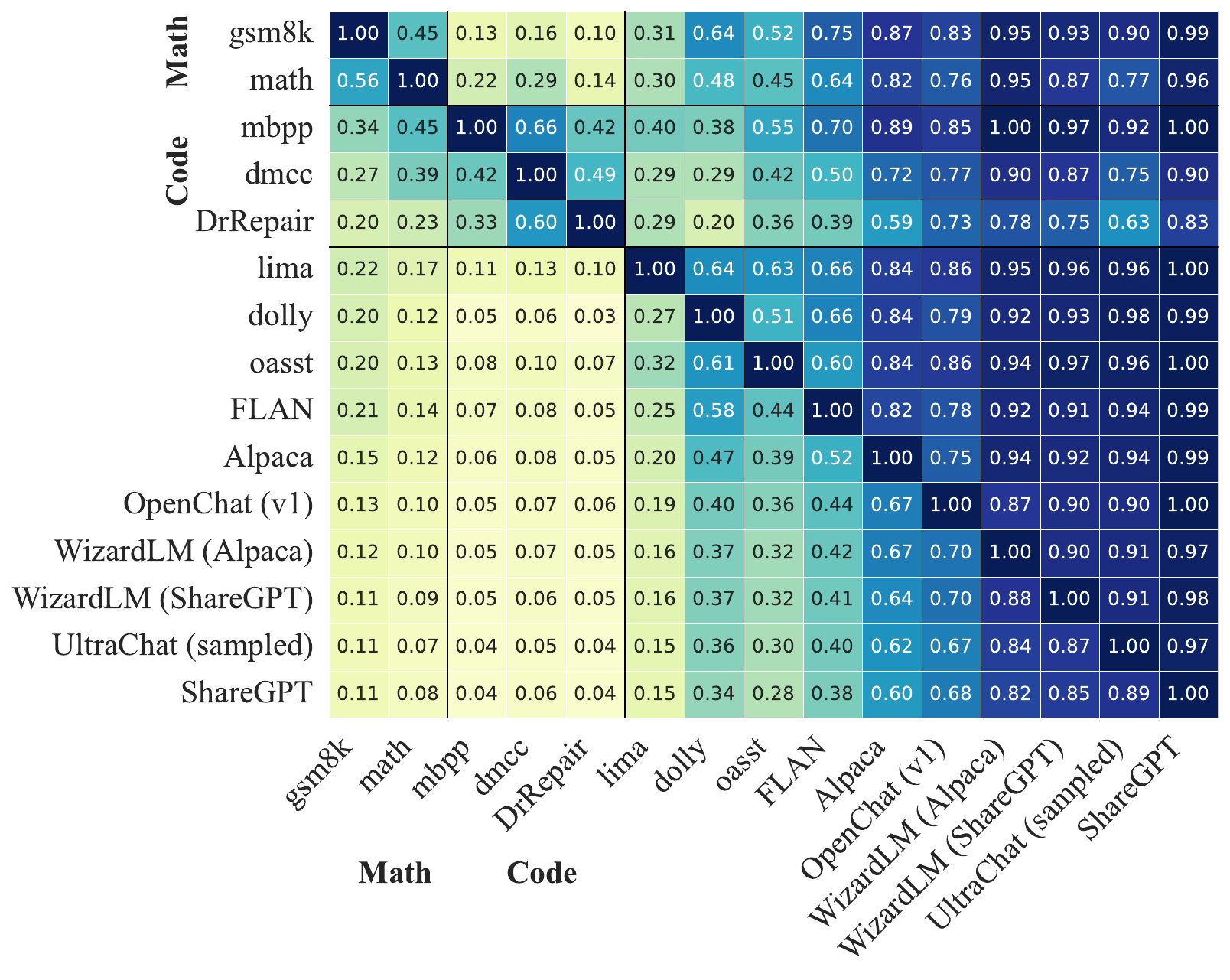}
        \label{subfig:data_correlation}
    }
    \caption{
    Open-source SFT dataset analysis based on tags.
    \Cref{subfig:diversity_and_complexity} shows diversities and complexities of various open-source SFT datasets based on metrics from tags, where datasets are marked in scatter sizes, and model performance on AlpacaEval~\citep{alpaca_eval} are marked in scatter colors.
    The AlpacaEval performance is collected from the official leaderboard and \cite{wang2023far}, datasets without AlpacaEval scores are marked in circles.
    \Cref{subfig:data_correlation} displays correlations among SFT datasets based on the recalls of tags.
    Annotations are the ratio using unique tags of the dataset in the column that recalls the dataset in the row.
    }
    \label{fig:opensource-dataset-analysis}
\end{figure}

\subsection{Preliminary Analysis}\label{sec:sft_data_analysis}

We present the preliminary analysis of existing open-source datasets through normalized tags in \Cref{fig:opensource-dataset-analysis}.
To start with, we introduce the diversity and complexity attributes of SFT datasets induced by our tagging result: 
\begin{itemize}[leftmargin=1em]
    \item \textbf{Diversity} is used to access the range of intentions and semantics covered by queries in a dataset. According to the tagging results, a dataset is considered more diverse if it covers more individual tags. The attribute is quantified as the unique tag coverage rate for the overall tag set. 
    \item \textbf{Complexity} aims to measure the number of intentions and semantics complicating queries. 
    We assume a more complex query  would be assigned more tags.
    The attribute is quantified as the average tag number assigned to queries in a dataset. 
\end{itemize}
We first depict the overall assessments of each dataset regarding the axis of diversity and complexity as shown in \Cref{subfig:diversity_and_complexity}. 
Each dataset is represented as a dot whose size indicates the dataset sample size, and color indicates the performance of LLMs fine-tuned thereon. As shown, 
(1) \textbf{The larger size, the more diverse.} On the diversity axis, the larger datasets contain human queries of higher diversity. The dataset for OpenChat-v1 is filtered from ShareGPT, resulting in high query diversity and complexity while having a relatively small size. 
(2) \textbf{The larger size, the more complex.} On the complexity axis, we can see that WizardLM (Alpaca) has a larger average tag number than the Alpaca dataset; hence is more complex. WizardLM (Alpaca) is created by complicating the queries from Alpaca datasets using Evol-Instrcut. This observation demonstrates that the average tag number can present the diversity of an SFT dataset. The complexity is also positively correlated with data size, except for mathematical reasoning and code generation. 
(3) \textbf{Math and Code show different trends.} The datasets for mathematical reasoning (MATH, GSM8K) and code generation (DMCC, MBPP, DrRepair) focus on specific downstream abilities and result in low diversity, while such datasets have relatively high complexity. 
(4) \textbf{More diverse and complex data induces higher performance.} ShareGPT, UltraChat, and OpenChat-v1 datasets lay at the upper-right corner of \Cref{subfig:diversity_and_complexity}, having both high diversity and complexity. Vicuna, UltraChat, and Openchat, which are respectively fine-tuned on the datasets, achieve cutting-edge performance among open-resourced models in aligning with human preference, as evaluated by public leaderboards (e.g., AlpacaEval \citep{alpaca_eval}). This scenario verifies that LLMs can benefit from fine-tuning more diverse and complex data for alignment. 

We demonstrate the correlations between datasets regarding unique tag recalls to understand the correlations between existing SFT datasets.
As depicted in \Cref{subfig:data_correlation}, we use the tag sets of the datasets on each column to calculate the recall with respect to the tag sets of the datasets on each row.
We can conclude from the figure that (1) \textbf{Tags can identify different tasks.} Datasets for mathematical reasoning and code generation tasks exhibit higher tag recalls within the tasks.
This demonstrates that the tags can identify the uniqueness of mathematical reasoning and code generation datasets compared to more general-purpose datasets.
(2) \textbf{One covers all.} WizardLM (Alpaca), WizardLM (ShareGPT), UltraChat, and ShareGPT, respectively, have higher tag recalls for other datasets.
This demonstrates that the four datasets contain more diversified queries and cover other datasets, consistent with the results in \Cref{subfig:diversity_and_complexity}.

Overall, \modelname provides a good tool for analyzing SFT datasets through the perspective of tagging. Existing SFT datasets differ in diversity and complexity as evaluated by the tagging results.

% \section{Experiments}
% \input{experiments}

\section{\modelname for Data Selection}
% \begin{wrapfigure}[20]{r}{0.55\textwidth}
% \makebox[0.55\textwidth][c]{
% \begin{minipage}[t]{0.55\textwidth}
% \centering
% \null 

% \end{minipage}
% }
% \end{wrapfigure}

As analyses shown in \Cref{sec:sft_data_analysis}, we notice fine-tuning LLMs on more diverse and complex datasets may benefit alignment performance.
Therefore, we present a data selection method supported by \modelname in this section and align LLMs with selected data to show the effectiveness of \modelname.
We introduced experimental setup~(\Cref{sec:experimental-setup}), results~(\Cref{sec:results}), and primary analyses related to query diversity and complexity~(\Cref{sec:analysis}).

\subsection{Experiment Setup}\label{sec:experimental-setup}
 
Based on the normalized tagging results and the preliminary analyses of existing datasets as presented in Figure \ref{fig:opensource-dataset-analysis}, we conduct fine-grained experiments to discuss the impact of data complexity and diversity through control variate methods. Under the correlation analyses in Figure \ref{subfig:data_correlation}, each dataset of WizardLM(Alpaca), WizardLM(ShareGPT), UltraChat, and ShareGPT maintains large tag recalls regarding other datasets. The four datasets also have the largest average tag numbers shown in Figure \ref{subfig:diversity_and_complexity}. These results indicate that the four datasets have high data diversity and complexity. Therefore, we pool the four datasets and create different subsets for delve-deep data complexity and diversity analysis.
The pooled dataset contains 306,044 samples with a tag set size of 6,398 and an average tag number of 4.48.

\begin{algorithm}[]
\caption{Complexity-first Diverse Sampling}\label{alg:sample}
\KwData{The Whole Pooled Dataset $\mathcal{D}$, Sub-Dataset Size $N$}
\KwResult{The Sampled Sub-Dataset $\mathcal{D}_s$}
Initialize Empty $\mathcal{D}_s$\;
Sorting Queries in $\mathcal{D}$ by tag number in descending\;
\While{$|\mathcal{D}_s| < N$}{
    Tag Set $\mathcal{T}_s^B\gets\emptyset$\;
    \ForEach{Query $q\in\mathcal{D}$}{
          \If{Query Tags $\mathcal{T}_q: |\mathcal{T}_s^B\cup\mathcal{T}_q|>|\mathcal{T}_s^B|$ }{
            $\mathcal{D}_s \gets \mathcal{D}_s \cup \{q\}$\;
            $\mathcal{T}_s^B \gets \mathcal{T}_s^B\cup\mathcal{T}_q$\;
            $\mathcal{D} \gets \mathcal{D} \setminus \{q\}$\;  
            \If{$|\mathcal{D}_s|$ equals to $N$}{
                Break\;
            } 
        }
  }
}
\Return{$\mathcal{D}_s$}
\end{algorithm}

% \begin{wrapfigure}[26]{r}{0.60\textwidth}
%     \centering
%     \includegraphics[width=0.60\textwidth]{fig/mtbench_radar.pdf}
%     \caption{Radar plot showing detailed scores of \lmname-13b-v1.0 and major baselines on eight subtasks of \textsc{MT-Bench}. Detailed numbers can be viewed in \Cref{tab:detailed_results}.}\label{fig:mtbench-radar}
% \end{wrapfigure}

In the following experiments, all the models fine-tuned are based on 13B version LLMs of either LLaMA \citep{touvron2023llama} or LLaMA-2 \citep{llama2}. If not specified otherwise, we fine-tune the model for five epochs with the batch size set to 128 and the learning rate set to $2\times10^{-5}$. The Vicuna-style template is applied to concatenate queries and responses during fine-tuning. 
We evaluate each fine-tuned model on \textsc{MT-Bench}\footnote{\url{https://huggingface.co/spaces/lmsys/mt-bench}} \citep{zheng2023judging} using GPT4 as a judge to demonstrate the alignment performance, set comparison to other LLMs, and conduct decoupled analyses on data complexity and diversity.

LLMs can benefit more from SFT datasets with higher diversity and complexity according to the analyses in \Cref{sec:sft_data_analysis}. We sample an SFT data subset of 6K samples from the pooled dataset with the highest sample complexity of an average tag number 16.56 and tag coverage of 100\%. We follow \Cref{alg:sample} to obtain the datasets.

\subsection{Results}\label{sec:results}

We use the dataset of 6K samples to align the 13B version of LLaMA \citep{touvron2023llama} and LLaMA-2 \citep{llama2} with human preference via SFT, and dub both aligned LLMs \lmname-13b-v1.0 and \lmname-13b-v2.0 respectively. We compare our models to two sets of baselines. We first use proprietary GPT-4 \citep{openai2023gpt4}, GPT-3.5 \footnote{\url{https://platform.openai.com/}}, and Claude-V1 \footnote{\url{https://www.anthropic.com/index/introducing-claude}} as strong state-of-the-art baselines, and then include strong cutting-edge open-resourced aligned LLMs, Vicuna \citep{vicuna2023}, WizardLM \citep{xu2023wizardlm}, Baize \citep{xu2023baize}, OpenChat \citep{openchat}, and Alpaca \citep{alpaca}. We leave the detailed introductions of these baselines to \Cref{app:model_baseline}. For a fair comparison, we all consider the open-resourced LLMs fine-tuned on the same 13B version of LLaMA. 

\begin{table}[h]
    \centering
    \small
    \setlength{\tabcolsep}{0.5mm}{
    \begin{tabular}{lcc}
    \toprule
    \textbf{Model} & \textbf{Data Size}& \textbf{MT-Bench} \\
    \midrule
    \multicolumn{3}{c}{\textbf{Proprietary Models}} \\
    \midrule
    gpt-4 & $-$ & 8.99 \\
    gpt-3.5-turbo & $-$  & $7.94$\\
    claude-v1 & $-$ & $7.90$ \\
    \midrule
    \multicolumn{3}{c}{\textbf{LLaMA-2 Based Open-source Models}} \\
    \midrule
    Llama-2-13b-chat \citep{llama2} &$-$ & $6.65$\\
    \lmname-13b-v2.0 & 6K & $6.55_{\pm 0.02}$ \\
    \midrule
    \multicolumn{3}{c}{\textbf{LLaMA Based Open-source Models}} \\
    \midrule
    alpaca-13b \citep{alpaca} & 52K & $4.53$ \\
    openchat-13b-v1 \citep{openchat} & 8K & $5.22$ \\
    baize-v2-13b \citep{xu2023baize} & 56K & $5.75$\\
    vicuna-13b-v1.1 \citep{vicuna2023} & 70K & $6.31$\\
    wizardlm-13b \citep{xu2023wizardlm} & 70K & $6.35$\\
    vicuna-13b-v1.3 \citep{vicuna2023} & 125K & $6.39$\\
    \lmname-13b-v1.0 & 6K & $\mathbf{6.44_{\pm 0.04}}$ \\
    \bottomrule
    \end{tabular}}
    \caption{
        Main results of \lmname.
        We present \textsc{MT-Bench} scores of both proprietary and open-source baselines in similar scales.
        We use the model name on the Huggingface Hub for all the open-sourced baselines. 
        We also report the approximate data size of each model used in supervised fine-tuning for alignment.
        We report the average of three \textsc{GPT-4} judgments and corresponding standard deviations and obtain results for other baselines from the official \textsc{MT-Bench} leaderboard.
        Dashes in the data column denote unknown data sizes.
        Detailed results are presented in \Cref{sec:detailed-results}.
    }
    \label{tab:maingories_results}
\end{table}

As shown in \Cref{tab:maingories_results}, we can see that \lmname-13b-v1.0 outperforms all the open-resourced aligned LLMs achieving a 6.44 average score on \textsc{MT-Bench}, although it is only fine-tuned based on LLaMA on only 6K samples which amount is far less than those of other LLMs.
We report the average of three \textsc{GPT-4} judgments as we notice there is noticeable randomness in \textsc{GPT-4} judgments.
We also provide the standard deviation of scores in three judgments.
This result illustrates that diversity and complexity do matter in improving human alignment performance via SFT. Our \modelname provides a decent tool for accessing and quantifying both attributes. \lmname-13b-v2.0 fine-tuned based on LLaMA-2 achieves even higher results while lagging behind LLaMA-2-chat by only 0.1 which is aligned with human preference via RLHF. When compared to proprietary high-performing LLMs, especially GPT-4, the performance is far lagging behind by 2.44 on \textsc{MT-Bench}.

\begin{figure}[]
\vspace{-0.5cm}
    \centering
    \includegraphics[width=0.60\textwidth]{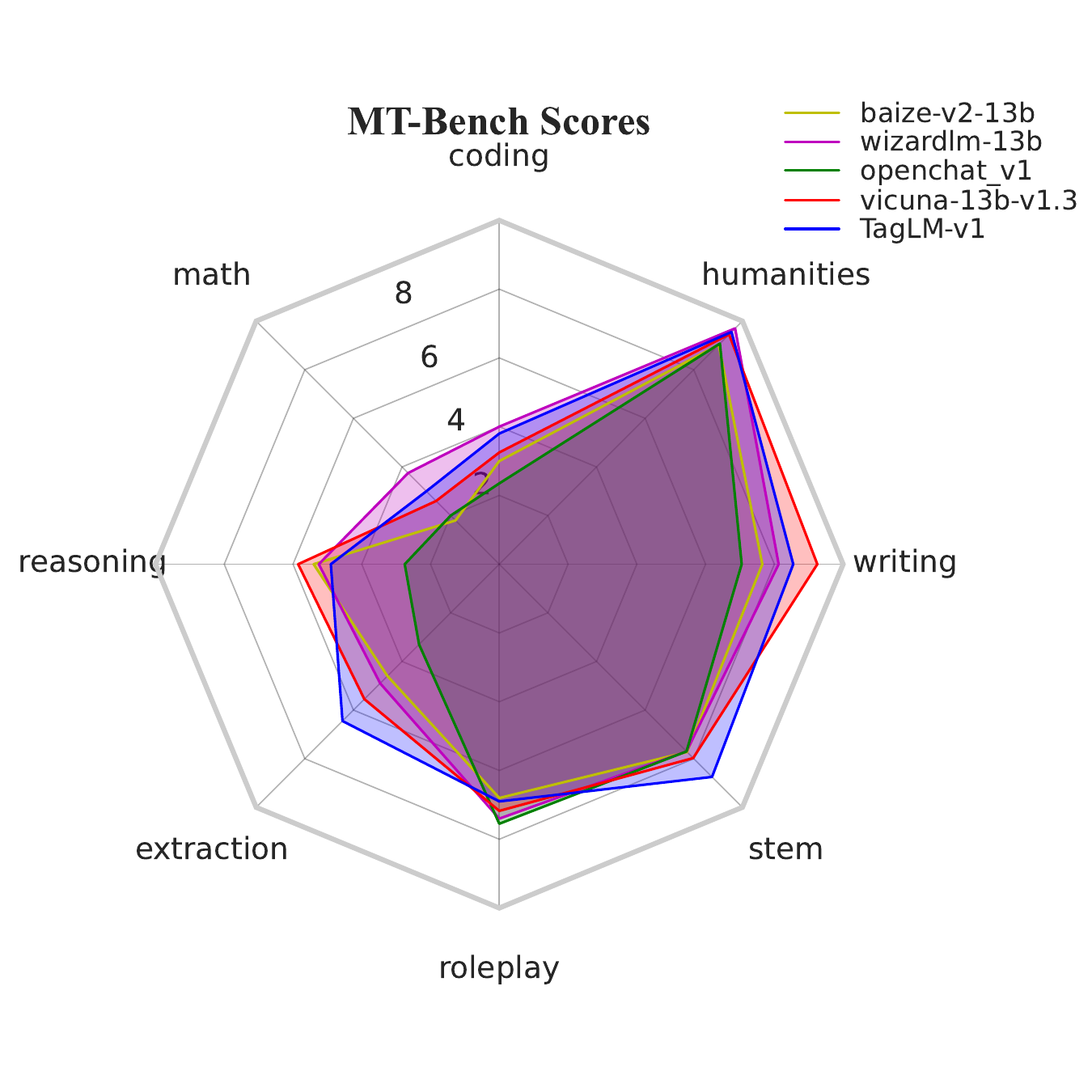}
    \caption{Radar plot showing detailed scores of \lmname-13b-v1.0 and major baselines on eight subtasks of \textsc{MT-Bench}. Detailed numbers can be viewed in \Cref{tab:detailed_results}.}\label{fig:mtbench-radar}
\end{figure}

We also present more detailed scores on \textsc{MT-Bench} in terms of eight tasks.
As shown in \Cref{fig:mtbench-radar}, \lmname-13b-v1 outperforms all other baselines on \textit{stem} and \textit{extraction}, and achieves comparable performances on \textit{humanities} with \textsc{Vicuna}, suggesting these tasks may rely on few data for alignment.
\lmname-13b-v1 ranks the second on \textit{math}, \textit{coding}, and \textit{writing}, but falls short on \textit{roleplay} and \textit{reasoning}.
These detailed results showing some tasks may require diverse but only a few alignment data, while tasks about reasoning and writing may continually benefit from large-scale data.

\begin{figure}[t]
    \centering
    \subfloat[Performance under different Tag Complexities]{
        \includegraphics[width=0.48\textwidth]{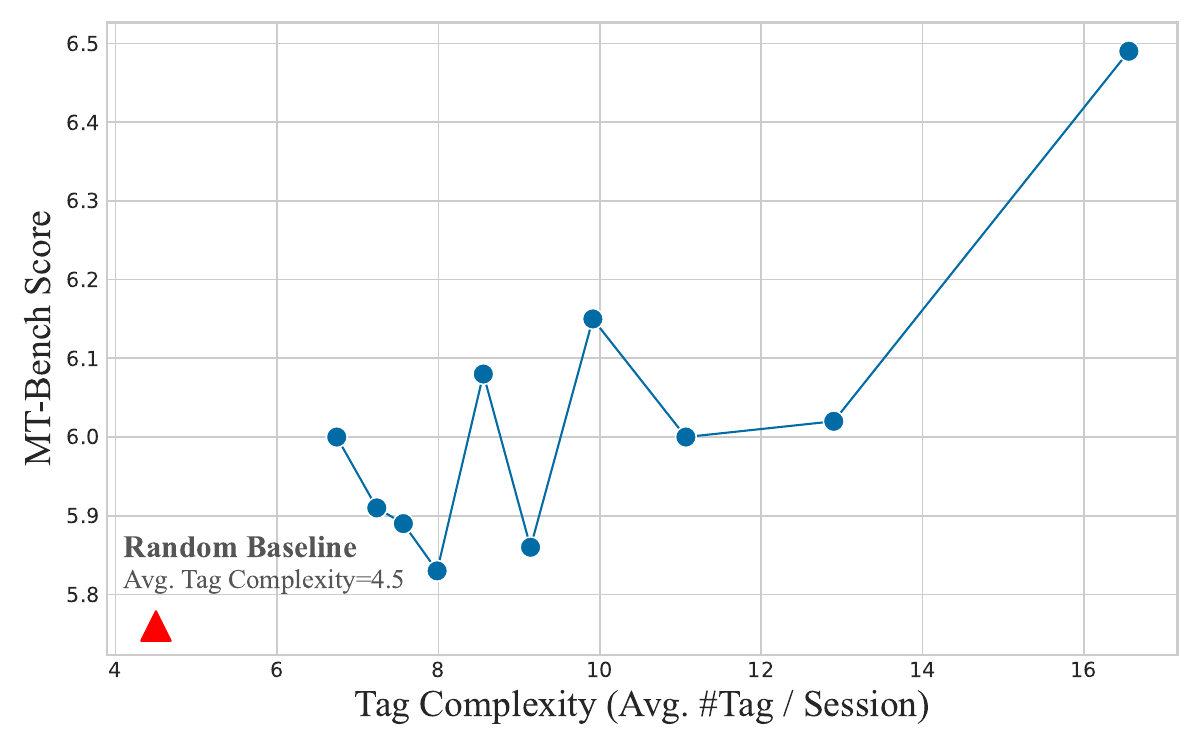}
        \label{subfig:result-complexity}
    }
    \subfloat[Performance under different Tag Diversities]{
        \includegraphics[width=0.48\textwidth]{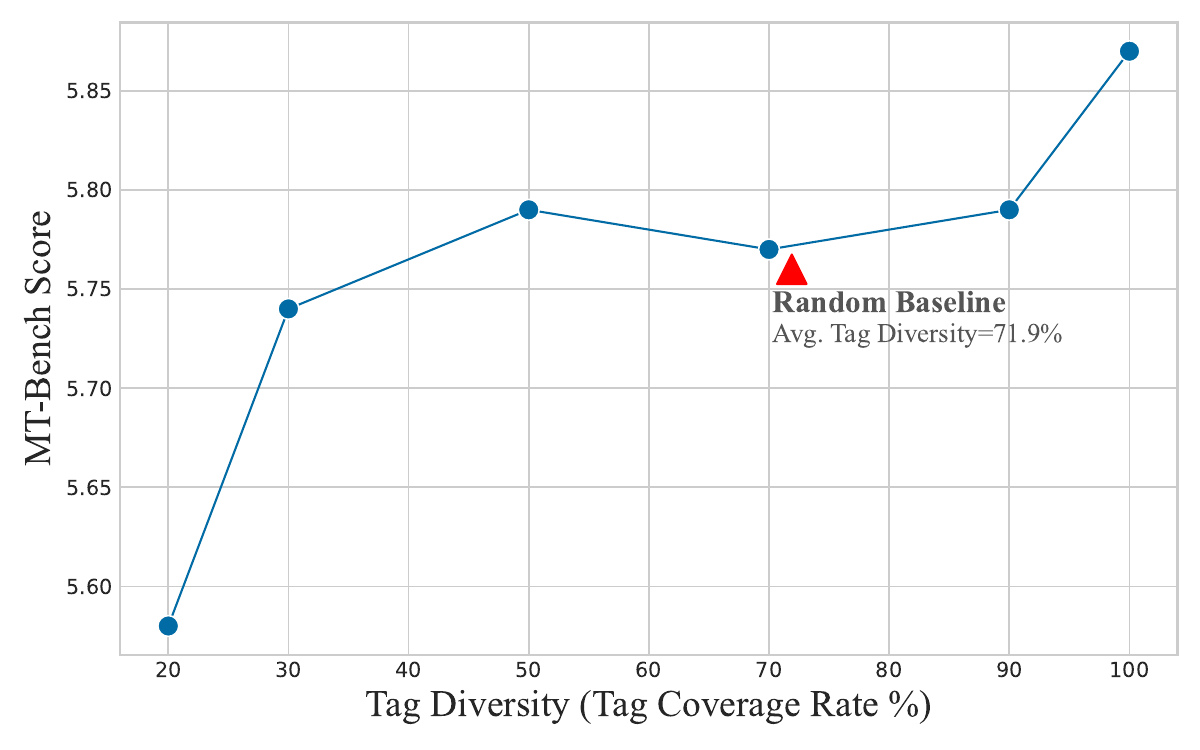}
        \label{subfig:result-diversity}
    }
    \caption{
    Analysis results of model performance in terms of different tag complexities and diversities.
    \Cref{subfig:result-complexity} shows \textsc{MT-Bench} scores over different tag complexities defined as an average number of tags per session.
    \Cref{subfig:result-diversity} shows scores over different tag diversities defined as coverage rates over all tags.
    We include a random baseline in both figures as shown in red triangles.
    }
    \label{fig:opensource-dataset-}
\end{figure}

\subsection{Decoupled Analysis}\label{sec:analysis}

We provide decoupled analyses of complexity and diversity to demonstrate how they influence alignment performance separately.

\xhdr{Complexity}
To decouple and focus on the data complexity, we sample different data subsets of diverse averaged tag numbers. Different sampled subsets share the same sample size of 6k and have the same tag coverage of 100\% which implies the largest data diversity. 
In the sampling procedure, all the data samples are first sorted by the tag numbers in descending order. Then for each data subset, we start from the sample in the whole dataset with the largest tag number. The sample that can expand the tag set size of the current sampled data will be extracted and removed from the whole dataset. 
If the tag set of the current sampled subset covers the whole tag set and the sample number is still less than 6k, we repeat the sampling procedure until the sample numbers reach 6k. This is a similar sampling procedure as \Cref{alg:sample}. We leave the detailed sampling algorithm for complexity analysis in \Cref{app:sampling}.

We sample 10 different data subsets, and the average tag numbers of the subsets range from 6.7 to 16.6. As a result shown in Figure \ref{subfig:result-complexity}, the overall trend of performance on \textsc{MT-Bench} is increasing along with the growth of average tag numbers. On the fine-grained level of average tag numbers where the number difference between subsets is small, this trend may not be significant. Compared to the randomly sampled datasets the average tag number of which is around 4.5, all the 10 data subsets can lead to superior fine-tuned model performance than the randomly sampled subset baseline. To summarize, on a coarse-grained level of data complexity the downstream performance is positively correlated to the average tag number, while on the fine-grained level, such a phenomenon becomes less evident. This may be partly because \textsc{ChatGPT} does not recall all the possible tags for each query or some tags are filtered out during the tag normalization procedure, resulting in a less accurate tag number for each query. 

\xhdr{Diversity}
For diversity, we sample different data subsets spanning various tag coverage rates regarding the whole tag set. Different subsets share the same sample scale of 6k and the same average tag number implying the same data complexity. The average tag number is set to 5.0.
For data subset sampling, we first draw samples that can expand the tag set size of the current sampled data until the target tag coverage rate. Then we traverse the remaining samples and extract samples that do not expand the tag coverage and can keep the current average tag number of the subset around 5.0. 
We leave the detailed sampling algorithm for diversity analysis in \Cref{app:sampling}.

We can observe in Figure \ref{subfig:result-diversity} that as the tag coverage increases the fine-tuned model can achieve higher \textsc{MT-Bench} scores. Randomly sampled data subsets of tag coverage 71.9\% result in similar model performance with the sampled subset of tag coverage 70\%. This demonstrates that through the scope of tags, the fine-tuned models may benefit from the more diverse datasets. The trend is not strictly linear and there seems a plateau ranging from 50\% to 90\% coverage. This could be due to the tags assigned may not share the sample importance for diversity, for example, the tags \textit{software development} may express more similar semantics with \textit{C++ programming} than those with \textit{information retrieval}.

% \xhdr{Sample Scale}

\section{Conclusion}
In this paper, we introduced \modelname, an open-set fine-grained tagger that leverages the instruction-following ability of modern chatbots like \textsc{ChatGPT}.
Tagging results on open-source SFT datasets show that aligning with more diverse and complex datasets may improve the performance of \modelname.
To explore this insight further, we conducted comprehensive scale analyses on the query diversity and complexity by sampling existing open-source data based on tags from \modelname.
We designed a complexity-first diverse sampling method to sample six thousand queries, and our LLMs fine-tuned on this selected dataset outperformed other open-source models aligned with considerably more data.
Moreover, further analyses revealed that model performance increases with more diversity and complexity.
In summary, our proposed \modelname provides a novel aspect for a deeper understanding of query distribution in the alignment of LLMs.
It has robust potential to be extended to more applications beyond the data selection shown in this work, such as creating comprehensive evaluations and tag-based self-instruct.

\section*{Limitations}
Our conclusions mainly rely on \textsc{MT-Bench} for model evaluations, which may miss some influence caused by SFT data.
Besides, we notice \textsc{MT-Bench} shows instabilities in terms of the randomness of \textsc{GPT-4} judgments, so we provide random ablations as comprehensive as possible to show the statistical significance of our results, including reporting standard variance of MT-Bench scores.
Furthermore, our analysis of SFT datasets is mainly focused on English, so our claims may not be directly extended to multi-lingual scenarios.

\bibliography{iclr2021_conference}
\bibliographystyle{iclr2021_conference}

\newpage
\appendix
\section*{Appendix}
\section{Tag Review}\label{app:sunburst}

\begin{figure}[t]
    \centering
    \includegraphics[width=\textwidth]{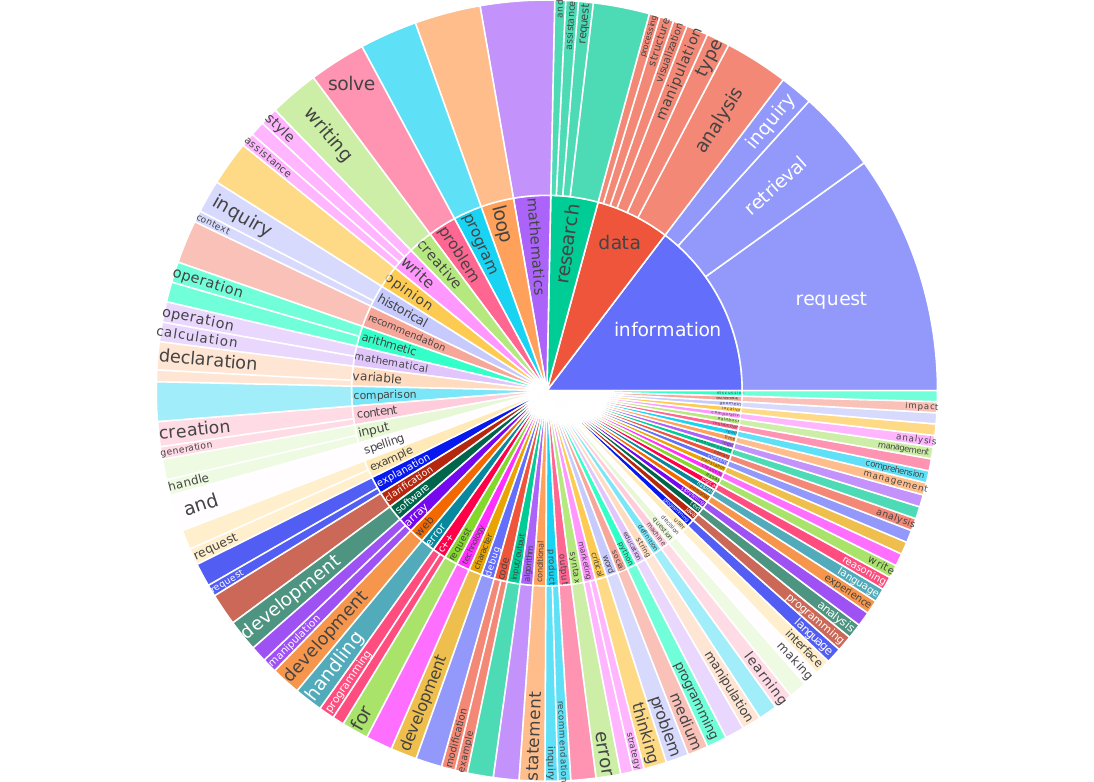}
    \caption{The sunburst plot of all tags.
    We plot with the first two words of each tag and the size is proportional to the frequency of the tag.}
    \label{fig:sunburst}
\end{figure}

We present a sunburst plot of all tags in \Cref{fig:sunburst} showing the most frequent tags is about information-related, data manipulations, and coding queries.
We plot with the first two words of each tag and the size is proportional to the frequency of the tag.
We only plot with tags that have frequencies larger than 2000 in our data pool.

\section{Prompt templates for \textsc{ChatGPT}}\label{sec:chatgpt-templates}

We preset our prompt for \textsc{ChatGPT} for annotation~(\Cref{tab:tagging_prompt}), precision evaluation~(\Cref{tab:precision_eval_prompt}), and consistency evaluation~(\Cref{tab:consistency_eval_prompt}).

\begin{table}[t]
    \centering
    \begin{tabular}{p{100mm}}
    \toprule
    You are a tagging system that provides useful tags for instruction intentions to distinguish instructions for a helpful AI assistant. Below is an instruction:\\
    
    [begin]
    
    \{instruction\}
    
    [end]
    
    Please provide coarse-grained tags, such as "Spelling and Grammar Check" and "Cosplay", to identify main intentions of above instruction.
    Your answer should be a list including titles of tags and a brief explanation of each tag.
    Your response have to strictly follow this JSON format: [\{"tag": str, "explanation": str\}].
    Please response in English. \\
    \bottomrule
    \end{tabular}
    \caption{\textsc{ChatGPT} prompt template for annotating intention tags of given queries.}
    \label{tab:tagging_prompt}
\end{table}

\begin{table}[t]
    \centering
    \begin{tabular}{p{100mm}}
    \toprule
    You are an experienced judge for intention tags of instructions. You will be provided a query and a list of tags describing intentions of the query as followed:

    [query]: \{query\}

    \{tags\}

    Please provide feedback about whether all tags precisely describe an intention of the instruction. Please identify all incorrect tags and provide their indices in the JSON format as your response.
    The JSON format for your response is a list of JSON dictionary and the JSON dictionary has only one key to identify the index of each incorrect tag: [\{"idx": int\}].
    For example, if [tag 0] and [tag 2] are incorrect, you should response as [\{"idx": 0\}, \{"idx", 2\}]. If all tags are correct, please response an empty list as [].\\
    \bottomrule
    \end{tabular}
    \caption{\textsc{GPT-4} prompt template for evaluating tagging precision.}
    \label{tab:precision_eval_prompt}
\end{table}

\begin{table}[t]
    \centering
    \begin{tabular}{p{100mm}}
    \toprule
    You are an experienced judge for consistency of intention tags for instructions. You will be provided a tag and a list of instructions labeled with this tag as followed:

    [tag]: \{tag\}

    \{instructions\}

    Please provide feedback about whether the meaning of this tag is consistent among all instructions. Please identify all inconsistent instructions and provide their indices in the JSON format as your response.
    The JSON format for your response is a list of JSON dictionary: [\{"idx": int\}].
    For example, if the meaning of tags in [instruction 0] and [instruction 2] are inconsistent, you should response as [\{"idx": 0\}, \{"idx": 2\}]. If the meaning of tag is consistent in all instructions, please response an empty list as [].\\
    \bottomrule
    \end{tabular}
    \caption{\textsc{GPT-4} prompt template for evaluating tagging consistency.}
    \label{tab:consistency_eval_prompt}
\end{table}

\section{Datasets}\label{sec:datasets}
We apply \modelname to 17 open-source SFT datasets for intention tagging:
\begin{itemize}[leftmargin=1em]
    \setlength\itemsep{-0.1em}
    \item \textbf{ShareGPT\footnote{Exact dataset of ShareGPT (\url{https://sharegpt.com/}) has not been released. We instead use a reproduced version from \url{https://huggingface.co/datasets/anon8231489123/ShareGPT_Vicuna_unfiltered/tree/main/HTML_cleaned_raw_dataset}, and follow Vicuna preprocess.}} refers to the multi-turn chatting histories used by \textsc{Vicuna}~\citep{vicuna2023}.
    ShareGPT includes human-written queries and responses from \textsc{ChatGPT} and other chatbots.
    \item \textbf{OpenChat}~\citep{openchat} is a subset of ShareGPT containing only chat histories with \textsc{GPT-4} responses.\footnote{We use the dataset with 8,000 GPT-4 responses denoting as OpenChat v1.0 in \url{https://huggingface.co/datasets/openchat/openchat_sharegpt4_dataset}}
    \item \textbf{UltraChat}~\citep{ding2023enhancing} is a systematically designed, diverse, informative, large-scale dataset of multi-turn instructional conversations without involving human queries.\footnote{\url{https://huggingface.co/datasets/stingning/ultrachat}}
    \item \textbf{Alpaca}~\citep{alpaca} is a dataset generated by the modified \textsc{Self-Instruct} method~\citep{wang2022self}, containing 52,000 instruction-following demonstrations generated from OpenAI’s \textit{text-davinci-003} model.\footnote{We collect the Alpaca dataset along with Dolly, OAssist, and Unnatural from the sharing of \cite{wang2023far}\url{https://github.com/allenai/open-instruct}.}
    \item \textbf{WizardLM}~\citep{xu2023wizardlm} is an instruction dataset built with the \textsc{Evol-Instruct} method.
    \textsc{Evol-Instruct} utilizes \textsc{ChatGPT} to augment the complexity of the same queries in Alpaca and ShareGPT.
    We denote these two subsets as WizardLM(Alpaca) and WizardLM(ShareGPT) for clarification.\footnote{We use the V2 version of WizardLM in \url{https://huggingface.co/datasets/WizardLM/WizardLM_evol_instruct_V2_196k}.}
    \item \textbf{FLAN}~\citep{weifinetuned} is a series of data from NLP tasks formatted in instruction tuning.
    The queries in FLAN are generated by templates for each NLP task.
    \item \textbf{Dolly}~\citep{DatabricksBlog2023DollyV2} contains 15,000 high-quality human-generated prompt and response pairs for instruction tuning of LLMs.
    \item \textbf{OAssist}~\citep{kopf2023openassistant} is a crowdsourced human-annotated dataset about multi-lingual conversations.
    \item \textbf{Unnatural}~\citep{honovich2022unnatural} contains queries generated by prompting \textsc{Davinci-002}.
    \item \textbf{Lima}~\citep{zhou2023lima} contains only 1,000 carefully human-curated prompts and responses.
    \item \textbf{Math Collections}: We involve a set of math datasets including GSM8K~\citep{cobbe2021training} and MATH~\citep{hendrycks2021measuring} to prompt \modelname generating fine-grained mathemetical tags.
    \item \textbf{Code Collections}: We also involve a set of code datasets including DMCC~\citep{dmcc}, MBPP~\citep{austin2021program}, and DrRepair~\citep{Yasunaga20DrRepair} for the same purpose as introducing mathematical datasets.
\end{itemize}

\section{Baseline LLMs}\label{app:model_baseline}

We give introductions to the LLM baselines for human alignment. 
\begin{itemize}[leftmargin=1em]
    \setlength\itemsep{-0.1em}
    \item \textbf{Alpaca} \citep{alpaca} is the first open-resourced LLM aligned with human preference. Alpaca is fine-tuned on SFT data of 52K samples generated from text-davince-003 using Self-Instruct \citep{wang2023selfinstruct}.
    \item \textbf{WizardLM} \citep{xu2023wizardlm} is fine-tuned on the SFT data enhanced with a novel technique named Evol-Instruct. It complexifies the Alpaca SFT data using \textsc{ChatGPT} and achieves better alignment performance. 
    \item \textbf{Vicuna} \citep{vicuna2023} is an aligned LLM fine-tuned on collected user chatting logs of proprietary high-performing chatbots on ShareGPT.
    \item \textbf{OpenChat} \citep{openchat} is fine-tuned on a subset of ShareGPT with only the chatting logs with GPT-4.
    \item \textbf{Baize} \citep{xu2023baize} uses 100K dialogues generated by self-chatting of \textsc{ChatGPT}. It also includes Alpaca's data for SFT. 
    \item \textbf{LLaMA-2 Chat} \citep{llama2} differs from the above-mentioned LLMs in (1) being based on per-trained LLaMA-2 instead of LLaMA \citep{touvron2023llama}; (2) being aligned with human preference by both SFT and RLHF.
\end{itemize}

\section{Sampling Algorithm For Decoupled Analysis}\label{app:sampling}

We present our sampling algorithm for decoupled analysis of complexity and diversity in \Cref{appalg:sample_complex}
 and \Cref{appalg:sample_diverse}, respectively.
 
\begin{algorithm}[t]
\caption{Data Sampling for Complexity Analysis}
\label{appalg:sample_complex}
\KwData{The Whole Pooled Dataset $\mathcal{D}$}, Subset Size $N$
\KwResult{The Sampled Sub-Dataset of Different Complexity $\mathrm{D}=\{\mathcal{D}_{c}^i|i=1,\dots,n\}$}
Sorting Queries in $\mathcal{D}$ by tag number in descending\;
Initialize $\mathrm{D}=\text{list}()$\;
\ForEach{$i$ in $\{1,\dots,n\}$}{
    Initialize Empty $\mathcal{D}_c^i$\;
    \While{$|\mathcal{D}_c^i| < N$}{
        Tag Set $\mathcal{T}_c^B\gets\emptyset$\;
        \ForEach{Query $q\in\mathcal{D}$}{
              \If{Query Tags $\mathcal{T}_q: |\mathcal{T}_c^B\cup\mathcal{T}_q|>|\mathcal{T}_c^B|$ }{
                $\mathcal{D}_c^i \gets \mathcal{D}_c^i \cup \{q\}$\;
                $\mathcal{T}_c^B \gets \mathcal{T}_c^B\cup\mathcal{T}_q$\;
                $\mathcal{D} \gets \mathcal{D} \setminus \{q\}$\;  
                \If{$|\mathcal{D}_c^i| = N$}{
                    $\mathrm{D} \gets\mathrm{D}$ appends $\mathcal{D}_c^i$\;
                    Break\;
                } 
            }
      }
    }
}
\Return{$\mathrm{D}$}
\end{algorithm}

\begin{algorithm}[t]
\caption{Data Sampling for Diversity Analysis}
\label{appalg:sample_diverse}
\KwData{The Whole Pooled Dataset $\mathcal{D}$, Preset Coverage Rate $\mathcal{R}=\{r^i|i=1,\dots,n\}$}, Subset Size $N$
\KwResult{The Sampled Sub-Dataset of Different Diversity $\mathrm{D}=\{\mathcal{D}_{d}^{r_i}|i=1,\dots,n\}$}

Initialize $\mathrm{D}=\text{list}()$\;
\ForEach{$i$ in $\{1,\dots,n\}$}{
    Initialize Empty $\mathcal{D}_{d}^{r_i}\gets\emptyset$\;
    Tag Set $\mathcal{T}_d\gets\emptyset$\;
    \ForEach{Query $q\in\mathcal{D}$}{
          \If{Query Tags $\mathcal{T}_q: |\mathcal{T}_d\cup\mathcal{T}_q|>|\mathcal{T}_d|$ }{
            $\mathcal{D}_{d}^{r_i} \gets \mathcal{D}_{d}^{r_i} \cup \{q\}$\;
            $\mathcal{T}_d \gets \mathcal{T}_d\cup\mathcal{T}_q$\;
            $\mathcal{D} \gets \mathcal{D} \setminus \{q\}$\;  
            \If{$|\mathcal{T}_d|/|\mathcal{T}| > r_i$}{
                Break\;
            } 
        }
    }
    \While{$|\mathcal{D}_{d}^{r_i}| < N$}{
        \ForEach{Query $q\in\mathcal{D}$}{
              \If{Query Tags $\mathcal{T}_q: |\mathcal{T}_d\cup\mathcal{T}_q|=|\mathcal{T}_d|$ }{
                $\mathcal{D}_{d}^{r_i} \gets \mathcal{D}_{d}^{r_i} \cup \{q\}$\;
                $\mathcal{D} \gets \mathcal{D} \setminus \{q\}$\;  
                \If{$|\mathcal{D}_{d}^{r_i}| = N$}{
                    Break\;
                } 
            }
      }
    }
    $\mathrm{D} \gets\mathrm{D}$ appends $\mathcal{D}_{d}^{r_i}$\;
}
\Return{$\mathrm{D}$}
\end{algorithm}

\section{Detailed Results}\label{sec:detailed-results}

We present our detailed results on \textsc{MT-Bench} in \Cref{tab:detailed_results}, which is also the source data of \Cref{fig:mtbench-radar}.

\begin{table}[t]
     \centering
     \small
     \setlength{\tabcolsep}{0.5mm}{
     \begin{tabular}{lc|cccccccc|cc}
     \toprule
     \multirow{2}{*}{\textbf{Model}} & \multirow{2}{*}{\textbf{Data}} & \multicolumn{8}{c}{\textbf{\textsc{MT-Bench} Scores}}\vline & \multicolumn{2}{c}{\textbf{Average}}\\
     & & code & extraction & humanities & math & reason & roleplay & stem & writing & all & w/o C\&M\\
     \midrule
     gpt-4 & $-$ & 8.55 & 9.38 & 9.95 & 6.8 & 9.0 & 8.9 & 9.7 & 9.65 & 8.99 & 9.43\\
     gpt-3.5-turbo & $-$ & 6.9 & 8.85 & 9.55 & 6.3 & 5.65 & 8.4 & 8.7 & 9.2 & 7.94 & 8.39\\
     claude-v1 & $-$ & 6.25 & 8.8 & 9.7 & 4.8 & 5.95 & 8.5 & 9.7 & 9.5 & 7.9 & 8.69\\
     \midrule
     Llama-2-13b-chat &-  & 3.0 & 6.92 & 9.75 & 3.45 & 5.1 & 7.5 & 8.62 & 8.85 & 6.65 & 7.79\\
     \lmname-13b-v2.0~(1) & 6K & 3.75 & 6.5 & 9.55 & 2.1 & 5.3 & 7.95 & 8.5 & 8.75 & 6.55 & 7.76\\
     \lmname-13b-v2.0~(2) & 6K & 3.7 & 6.2 & 9.52 & 2.15 & 5.35 & 8.1 & 8.4 & 8.95 & 6.55 & 7.75\\
     \lmname-13b-v2.0~(3) & 6K & 3.4 & 7.35 & 9.6 & 2.15 & 5.9 & 7.45 & 8.28 & 8.0 & 6.52 & 7.76\\
     \midrule
     \modelname-v1.0-13b~(1) & 6K & 3.8 & 6.45 & 9.55 & 3.0 & 4.9 & 6.9 & 8.75 & 8.55 & 6.49 & 7.52\\
     \modelname-v1.0-13b~(2) & 6K & 3.45 & 6.35 & 9.65 & 2.95 & 4.95 & 7.15 & 8.65 & 8.5 & 6.46 & 7.54\\
     \modelname-v1.0-13b~(3) & 6K & 3.4 & 6.45 & 9.45 & 2.85 & 5.05 & 7.05 & 8.43 & 8.4 & 6.38 & 7.47\\
     vicuna-13b-v1.3 & 125K & 3.25 & 5.55 & 9.45 & 2.6 & 5.85 & 7.18 & 7.98 & 9.25 & 6.39 & 7.54\\
     vicuna-13b-v1.1 & 70K & 2.95 & 6.4 & 9.45 & 2.9 & 4.65 & 7.5 & 8.55 & 8.05 & 6.31 & 7.43\\
     wizardlm-13b & 70K & 4.0 & 4.9 & 9.7 & 3.75 & 5.25 & 7.4 & 7.7 & 8.12 & 6.35 & 7.18\\
     baize-v2-13b & 56K & 3.0 & 4.6 & 9.02 & 1.8 & 5.4 & 6.8 & 7.72 & 7.65 & 5.75 & 6.87\\
     nous-hermes-13b & 300K & 2.45 & 5.05 & 9.0 & 2.65 & 3.8 & 6.38 & 7.02 & 7.75 & 5.51 & 6.5\\
     gpt4all-13b-snoozy & 900K & 3.0 & 4.8 & 8.85 & 1.2 & 4.2 & 7.0 & 6.9 & 7.35 & 5.41 & 6.52\\
     koala-13b & 472K & 2.9 & 4.15 & 8.45 & 1.9 & 4.0 & 6.85 & 7.2 & 7.35 & 5.35 & 6.33\\
     openchat-13b-v1 & 8K & 2.35 & 3.3 & 9.07 & 2.0 & 2.75 & 7.55 & 7.7 & 7.05 & 5.22 & 6.24\\
     alpaca-13b & 52K & 2.35 & 4.15 & 7.85 & 1.05 & 3.5 & 5.45 & 5.2 & 6.7 & 4.53 & 5.48\\
     \bottomrule
     \end{tabular}}
     \caption{
         Main results of \modelname.
         We present \textsc{MT-Bench} scores of both proprietary and open-source baselines in similar scales.
         We also provide average scores overall categories and categories without code and math~(w/o C\&M).
         Dashes in the data column denote unknown data scales.
         Parentheses mark the three different rounds of \textsc{GPT-4} judgments.
     }  
    \label{tab:detailed_results}
\end{table}

\section{Counterfactual Evaluation}

To test how well annotators can evaluate tag quality, we created counterfactual cases for two tasks.
In the tag precision task, we substituted some tags with similar ones in terms of semantics.
In the tag consistency task, we used inconsistent instructions to replace the original instructions.
Both humans and \textsc{GPT-4} are able to recognize most of the counterfactual cases.
And humans are better at tag precision, while \textsc{GPT-4} is better at tag consistency.
This analysis shows that annotators have low false positive rates and proof confidence of their judgments in the original tagging results.

\begin{table}[t]
    \centering
    \small
    \setlength{\tabcolsep}{1mm}{
    \begin{tabular}{ccccc}
    \toprule
    \multirow{2}{*}{\textbf{Metric}} & \multicolumn{2}{c}{\textbf{GPT-4 Annotation}} & \multicolumn{2}{c}{\textbf{Human Annotation~(1\%)}}\\
    & Original & Counterfactual & Original & Counterfactual \\
    \midrule
    \textbf{Tag Precision} & 96.1 & 6.1 & 100 & 0\\
    \textbf{Tag Consistency} & 86.6 & 7.8 & 100 & 14.3\\
    \bottomrule
    \end{tabular}}
    \caption{
    Evaluation for the tagging quality of \modelname.
    We design two metrics, tagging precision and consistency, for evaluating \modelname.
    We employ \textsc{GPT-4} to label 4,000 tagging results.
    And we also employ three human annotators to annotate 1\% cases and report their \underline{majority voting}.
    We also create counterfactual cases to probe the judgment abilities of different annotators.
    }
    \label{tab:tagging_evaluation_counterfactual}
\end{table}

\section{Case from \modelname}

We present a sample case in \Cref{apptag:case1} to illustrate the behavior of \modelname.

\begin{table}[t]
    \centering
    \begin{tabular}{p{35mm}p{100mm}}
    \toprule
    \multicolumn{2}{c}{\textbf{Sample Case}} \\
    \midrule
    \multicolumn{2}{c}{Utterance} \\
    Human: & List 3 breakfast foods that are healthy.\\
    LLM:  & 1.Oatmeal - it's high in fiber, protein, and antioxidants, and can lower cholesterol levels.\\
    &2.Greek yogurt - it's packed with protein, calcium, and probiotics, which support gut health.\\
    &3.Eggs - they're rich in protein, vitamins D and B12, and minerals like selenium, and can help you feel full longer."\\
    \midrule 
    \multicolumn{2}{c}{\modelname Results} \\
    \multicolumn{1}{c}{\textbf{Tag}} & \multicolumn{1}{c}{\textbf{Explanation}} \\
    food recommendation & The instruction asks for a list of breakfast foods that are healthy.\\
    quantity & The instruction specifies that the list should include three breakfast foods.\\
    health & The instruction emphasizes the need for the breakfast foods to be healthy.\\
    \bottomrule
    \end{tabular}
    \caption{One case from SFT data of WizardLM tagged by \modelname. \modelname accurately captures the semantics (food, health) and intentions (recommendation, quantity) in the human query. }
    \label{apptag:case1}
\end{table}

\section{\textsc{InsTagger}: Local Tagger by Distillation}\label{app:tagger}

When it comes to localized deployments or large-scale tagging demands in which cases using \textsc{ChatGPT} or GPT-4 through OpenAI API is not available or infeasible, we may alter to a specialized smaller LLM in the aid of the scenarios. 
Therefore, we proposed \textsc{InsTagger}, which is equipped with the tagging ability of these high-performing chatbots by distillation. 
Distilling is an effective method to inject a smaller model with specialized abilities, which has been applied to mathematical reasoning abilities recently \citep{fu2023specializing}. We use our \modelname results on open-resourced SFT datasets to fine-tune a 7B version LLaMA-2 model. We use a template:

"\textit{You are a helpful assistant. Please identify tags of user intentions in the following user query and provide an explanation for each tag. Please respond in the JSON format \{"tag": str, "explanation": str\}. Query: $<$query-to-tag$>$ Assistant: $<$tagging-results$>$}"

to concatenate queries to tag and tagging results, and we also include the explanation in the tagging results to make the fine-tuned model obtain better tagging performance. The overall sample size for fine-tuning is 773,511 where we randomly sample 1,000 samples for validation. The model is fine-tuned with 512 batch size for 1 epoch since we empirically find that training for more than 1 epoch will lead to over-fitting. 

We validate the model on our validation set.
The tag-level F1 score based on exact match (EM) and semantic-based fuzzy match are 31.8\% and 73.4\%.
As this is an unconstrained open-generated tagging, EM is a very strict metric for annotating over six thousand tags.
Therefore, we also calculate the fuzzy match by \textsc{PhraseBERT}, which considers a predicted tag is correct if it has over 0.8 cosine similarity in semantics with any gold tag.

\end{document}